\documentclass[10pt,twocolumn,letterpaper]{article}

\usepackage{wacv}
\usepackage{times}
\usepackage{epsfig}
\usepackage{graphicx}
\usepackage{amsmath}
\usepackage{amssymb}

\usepackage{graphicx}
\usepackage{booktabs}
\usepackage{balance}

\def \shape{\mathbf{z}}
\def \pose{\boldsymbol{\xi}}

\def \points{\mathcal{X}}

\def \meanshape{\boldsymbol{\mu_{\widetilde{\phi}}}}

\newcommand{\refeq}[1]{Equation~\ref{eq:#1}}
\newcommand{\reffig}[1]{Figure~\ref{fig:#1}}
\newcommand{\reftab}[1]{Table~\ref{tab:#1}}

\graphicspath{{./images/}}

\usepackage[pagebackref=true,breaklinks=true,letterpaper=true,colorlinks,bookmarks=false]{hyperref}

\wacvfinalcopy

\ifwacvfinal\fi
\begin{document}

\title{SAMP: Shape and Motion Priors for 4D Vehicle Reconstruction}

\author{Francis Engelmann \hspace{2cm} J\"org St\"uckler \hspace{2cm} Bastian Leibe \\
Computer Vision Group, Visual Computing Institute, RWTH Aachen University\\
{\tt\small \{engelmann,stueckler,leibe\}@vision.rwth-aachen.de}
}

\maketitle
\ifwacvfinal\thispagestyle{empty}\fi

\begin{abstract}
Inferring the pose and shape of vehicles in 3D from a movable platform still remains a challenging task due to the projective sensing principle of cameras,
difficult surface properties e.g. reflections or transparency, and illumination changes between images.
In this paper, we propose to use 3D shape and motion priors to regularize the estimation of the trajectory and the shape of vehicles in sequences of stereo images.
We represent shapes by 3D signed distance functions and embed them in a low-dimensional manifold.
Our optimization method allows for imposing a common shape across all image observations along an object track.
We employ a motion model to regularize the trajectory to plausible object motions.
We evaluate our method on the KITTI dataset and show state-of-the-art results in terms of shape reconstruction and pose estimation accuracy.
\end{abstract}

\section{Introduction}
In recent years, vision-based approaches to scene understanding in urban street scenes have gained increasing attention in research on autonomous vehicles.
In contrast to LiDAR or radar, cameras are inexpensive, light-weight and passive sensors.
The processing of visual information to robustly track dynamic objects such as cars and to accurately estimate their 3D trajectory and shape is a challenge.
In general, extracting 3D information from images is an ill-posed problem due to the passive projective sensing principle.
Surface properties such as reflectiveness or transparency, as well as illumination changes, make it hard to establish accurate and reliable pixel correspondences between images to perform structure-from-motion.

In this paper, we propose to use 3D shape and object motion priors to better condition the problem of 3D shape and trajectory estimation of objects from image sequences.
Our approach builds upon a standard tracking-by-detection method but can be integrated with any other state-of-the-art tracker.
The tracker links 3D object detections (3DOP~\cite{Chen2015_3DOP}) between frames which we use to segment stereo depth observations on objects.
We represent vehicle shapes as 3D signed distance functions (SDFs) and embed them using a linear subspace.
This low-dimensional shape representation allows us to jointly optimize for the trajectory and shape parameters of a vehicle throughout its track.
Importantly, the shape estimate is shared across all the stereo observations along a given track.
Our method also imposes a motion model as a prior on the feasible behavior of the objects.
By this, our approach recovers the 3D object shape that best explains the stereo depth observations across all frames and couples distant but noisy depth observations with more accurate close-by observations along a trajectory.

\begin{figure}[t]
\centering
\includegraphics[trim=0 15 0 -15, clip,width=1.0\linewidth]{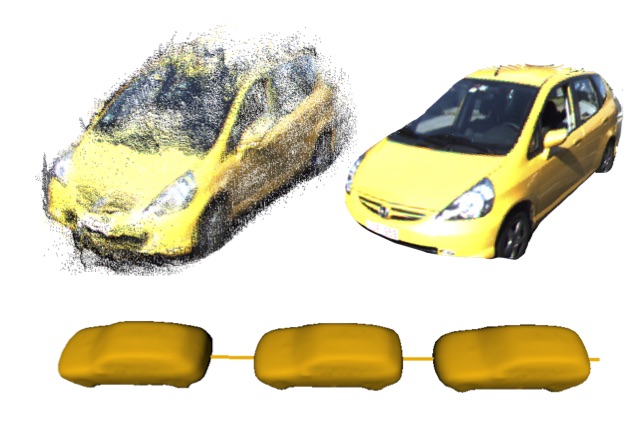}
\caption{\textbf{Results of our Method.} From a sequence of noisy stereo depth observations, our method jointly estimates the shape and motion of tracked vehicles by using a shape manifold and a motion model. Top left:  Stereo depth observations superimposed based on their estimated poses. Top right: Color image projected onto estimated shape. Bottom: Estimated shape and trajectory over time.}
\label{fig:teaser}
\end{figure}

We evaluate our approach on the KITTI dataset and demonstrate superior results in shape reconstruction and object pose estimation towards state-of-the-art baseline methods.
Our method boosts shape estimation accuracy for a tracked object and significantly improves the individual object poses along the track.

In summary, we make the following contributions:
(1) We propose a vision-based method for jointly estimating the 3D shape and the 3D trajectory of vehicles in urban street scenes.
Our method is formulated as an optimization problem regularized by shape as well as motion priors.
(2) We present a method to align 3D observations by the intermediate use of a shape embedding. This embedding serves as a shape prior shared among all detections along a track.
(3) We concurrently impose a motion model on the tracked objects which regularizes the estimated trajectories to smooth and typical object motions.

\section{Related Work}
Several methods have been proposed that fit CAD models or shape priors to single images in order to recover the 3D pose or shape of objects.
Zia et al.~\cite{ZiaPAMI13} use a PCA embedding of a wireframe model. The method parametrizes the wireframe in the location of object parts which are detected and localized in the image in order to retrieve the shape and pose of the object.
In own prior work ~\cite{EngelmannGCPR16} we use 3DOP to detect vehicles in stereo images and estimate their pose coarsely. There we fit truncated SDF shape priors of vehicles into the stereo depth reconstructions of the single stereo frames.
In this work, we track objects using a tracking-by-detection approach and performs multi-frame alignment using a shared shape prior.
Additionally, we impose a motion model on the 3D object trajectories to further regularize the joint estimate with a plausible motion prior.
Also related to our method are approaches that fit human body shape models into range scans such as SCAPE~\cite{anguelov2005_scape}.
SCAPE finds PCA embeddings of human body wireframe models and relies on annotated markers to complete single partial range scans.

In recent years, also object-based alignment over multiple frames has been investigated for scene understanding in urban street scenes.
Feni et al.~\cite{FenWACV2015} estimate the yaw angle of vehicles from annotated 2D detections.
Menze et al.~\cite{MenzeCVPR15} model moving rigid objects for improving stereo scene flow estimation between two consecutive frames.
However, they do not align an explicit shape model of objects, while our method provides object trajectories and shape over an arbitrary number of frames.
Song~\cite{SongCVPR15} couples object detection and monocular SfM in order to achieve object tracking and 3D pose estimation.
The full 3D object shape is not reconstructed in the process.
Ortiz-Cayon et al.~\cite{Ortizcayon3DV16} match 3D CAD models of objects into multi-view object detections.
They align 3D shape and silhouette of the model with the multi-view 3D reconstruction and 2D images, respectively.
Using 3D laser scanners, Held et al.~\cite{held2015_lasertracker} propose a probabilistic formulation and an anytime algorithm to register 3D scan segments on object over time combining shape, color and motion cues.
We use shape and motion priors to regularize the dense registration of stereo depth observations,
and we share the estimated shape across all frames.

Related to our approach are SLAM methods that explicitly include objects into the state-space of the SLAM optimization.
Dame et al.~\cite{DameCVPR13} model objects using GP-LVM shape priors and use them to regularize dense depth reconstruction in monocular SLAM in table-top scenarios.
In SLAM++~\cite{salasmoreno2013_slampp} object instances are detected and their 6-DoF pose is estimated in RGB-D images based on 3D CAD models of the objects.
The object pose estimates are included as spatial constraints between the camera pose and the objects in the SLAM graph optimization.
Ma et al.~\cite{lingni16icra} include planar objects into the SLAM state for RGB-D SLAM.
They also track camera motion through concurrent direct image alignment towards keyframes and model planes.
These methods assume the objects to remain static during the SLAM process.
Our method allows for several moving objects in the scene and optimizes for their motion and shape simultaneously.

\begin{figure*}[!tb]
\centering
  \includegraphics[trim=0 0 0 0, clip,width=1.0\linewidth]{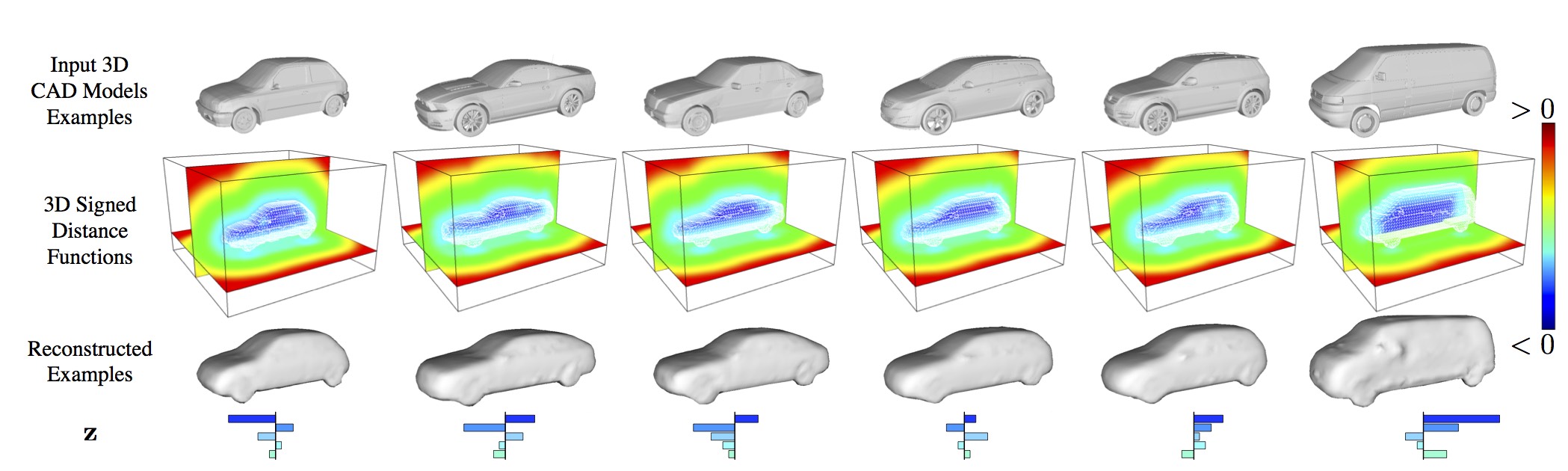}
  \caption{\textbf{Visualization of Shape Model.} First row: Input CAD models used to learn the low dimensional shape manifold. Second row: For each CAD model, we compute its corresponding 3D signed distance function $\widetilde{\phi}$. Positive distances are shown in red, negative ones in blue. The zero-level set is shown as a white wireframe.
  Third row: Reconstructed input models from our low-dimensional manifold. The bar-plots in the last row visualize the  value of the five dimensional shape encoding $\shape$. The top blue bar corresponds to the coefficient of the eigenvector with the largest eigenvalue.}
\label{fig:shape_model}
\end{figure*}

\section{Method}
Our method builds on several methods for stereo reconstruction, object detection and tracking.
We first perform stereo reconstruction using off-the-shelf algorithms~\cite{GeigerACCV10,YamaguchiECCV14} in all stereo image pairs individually.
We then detect vehicles using 3DOP~\cite{Chen2015_3DOP} and associate the detections into tracks using the approach of Geiger et al.~\cite{GeigerPAMI14}.
Additionally, the ego-motion of the observing camera is estimated using the visual odometry of Geiger et al.~\cite{GeigerIV11}
which allows us to transform stereo depth observations and detections into a common world coordinate system for all further inference steps.
At this stage, we obtained a set of tracks of varying length, each one consisting of multiple 3D detections.
Our method proceeds to estimate the shape of each tracked object and to refine the poses of the individual 3D detections associated to a track.
For each track, we impose a common shape and a motion model that jointly bind together the individual observations over time.
We formulate this problem using a probabilistic model which leads to an energy function that can be efficiently optimized.
Next, we describe the different components of our method in more detail.

\subsection{Shape Manifold}
Similar to our approach in~\cite{EngelmannGCPR16}, we learn a shape manifold by applying PCA to 3D CAD models from Google Warehouse. 
We bring the models to metric scale and align them at their centers with a common orientation (see \reffig{shape_model}). By relying on SDFs for representing shape representations, dense surface correspondences between models are not required.
We compute the SDF of a CAD model as follows: First, we sample depth observations from the surface of the CAD model.
We extract a point cloud of the object by rendering the CAD model from multiple view points.
Next, we approximate the continuous signed distance function in a discrete 3D grid.
In each voxel of the grid, we store the distance to the closest point from the object point cloud.

We briefly introduce the notation of the PCA shape embedding.
The continuous SDF~$\phi( \mathbf{x}, \mathbf{z} )$ provides a mapping from 3D points~$\mathbf{x}$ to the closest distance to the object surface.
The SDF is also parametrized in the shape encoding~$\mathbf{z} \in \mathbb{R}^R$.
The object surface corresponds to the zero level set of the SDF.
When discretizing the SDF in a 3D grid, we retrieve an approximation to the continuous SDF at an arbitrary position $\textbf{x}$ through trilinear interpolation of the grid SDF values~$\widetilde{\phi}_i( \mathbf{z} )$.
The grid SDF values correspond to the evaluated continuous SDF at the grid positions $i$, but still depend on the shape encoding~$\mathbf{z}$.
In order to perform the subspace embedding of the SDF samples from the CAD models using PCA, we stack the grid SDF values of each sample in a vector~$\mathbf{\widetilde{\phi}}$ and determine the sample covariance~$\mathbf{\Sigma}$ of the discretized SDFs.
We determine the subspace projection and backprojection
\begin{align}
   \shape\left( \boldsymbol{\widetilde{\phi}} \right) &= \mathbf{W}^{\top} \left( \boldsymbol{\widetilde{\phi}} - \meanshape \right),\\
   \boldsymbol{\widetilde{\phi}}( \mathbf{z} ) &= \mathbf{W} \, \shape + \meanshape,
\end{align}
through eigen-decomposition of the sample covariance, i.e. $\mathbf{\Sigma} = \mathbf{V} \mathbf{D} \mathbf{V}^{\top}$ and restricting~$\mathbf{W}$ to the eigenvectors corresponding to the~$R$ largest eigenvalues in~$\mathbf{D}$.
We denote the mean of the grid SDF vectors by $\meanshape$.

\subsection{Probabilistic Model}
\begingroup
\setlength{\thickmuskip}{3mu}
In the following, we describe the probabilistic model with which we explain the stereo depth observations of a tracked vehicle.
Its graphical model is depicted in \reffig{graphical_model} and shows which variables are probabilistically independent in our approach.
A track spans over~$T$ time-sequential stereo frames.
We denote the set of poses and depth observations of a track by $\pose = \{ \pose_1, ... , \pose_T \} $ and $\points = \{ \points_1, ... , \points_T \}$, respectively.
The pose $\pose_t \in \mathbb{R}^6$ of a detection at time $t$ is parametrized by a translation $\textbf{t} \in \mathbb{R}^3$, a rotation $\theta_y \in \mathbb{R}$ along the gravity axis, a translational velocity $v \in \mathbb{R}$ in heading direction and a rotational velocity $\omega \in \mathbb{R}$ along the vertical axis.
An observation $\points_t$ represents a set of stereo depth observations originating from the surface of a detected vehicle at time $t$ i.e. a set of points in 3D-space.
We assume the underlying shape $\shape$ of a tracked vehicle is constant for all observations of a given track.

\begin{figure}[t]
\centering
\includegraphics[width=0.9\linewidth]{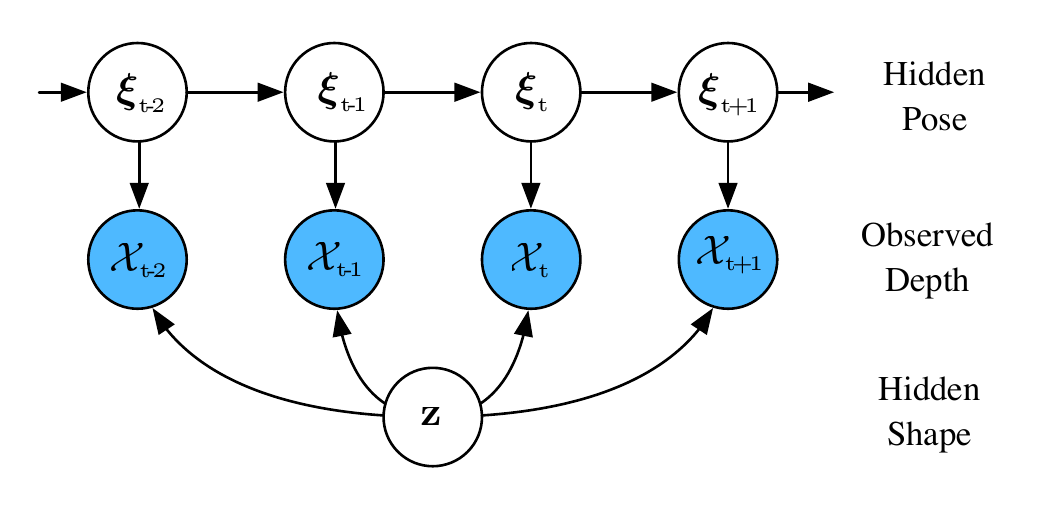}
\caption{\textbf{Graphical Model.} Each tracked object has a shape $\shape$. At each time step $t$, we obtain stereo depth observations $\points_t$ of the objects surface from a different viewpoint. A common motion model links together the poses $\pose$ of the object.}
\label{fig:graphical_model}
\end{figure}

Given the depth observations $\points$, we want to infer the shape $\shape$ and the poses $\pose$ of a tracked vehicle.
Hence, our method seeks to optimize the posterior of shape and pose estimates in a track given the stereo depth observations,
\begin{equation}\label{eq:posterior}
     p( \shape , \pose \mid \points ) = \eta \, p(\points \mid \shape , \pose) \, p(\shape) \, p(\pose),
\end{equation}
where $\eta$ is a normalization constant.
In this scenario, it is reasonable to assume that $p(\shape,\pose) = p (\shape) \, p(\pose)$, i.e. the shape of a vehicle is independent of its pose.

The posterior in \refeq{posterior} can be further factorized to
\begin{align}
	p(\pose) &= p(\pose_1) \underset{t=2}{\overset{T}{\prod}}p(\pose_{t} \mid \pose_{t-1}),\\
	 p( \points \mid \shape , \pose ) &= \underset{t=1}{\overset{T}{\prod}} \, \underset{\mathbf{x} \in \points_t}{\prod}p(\mathbf{x} \mid \shape, \pose_t),
\end{align}
where $\mathbf{x} \in \mathbb{R}^3$ is the 3D point corresponding to a stereo depth observation.
We find the optimal parameters by maximizing the posterior probability
\begin{equation}
	\{\shape, \pose \}^* = \underset{\shape, \pose}{\text{argmax}} \, p(\shape, \pose \mid \points).
\end{equation}
\endgroup

\subsection{Cost Function}
\begin{figure}[tb]
\centering
\includegraphics[trim=0 0 0 0, clip,width=1.0\linewidth]{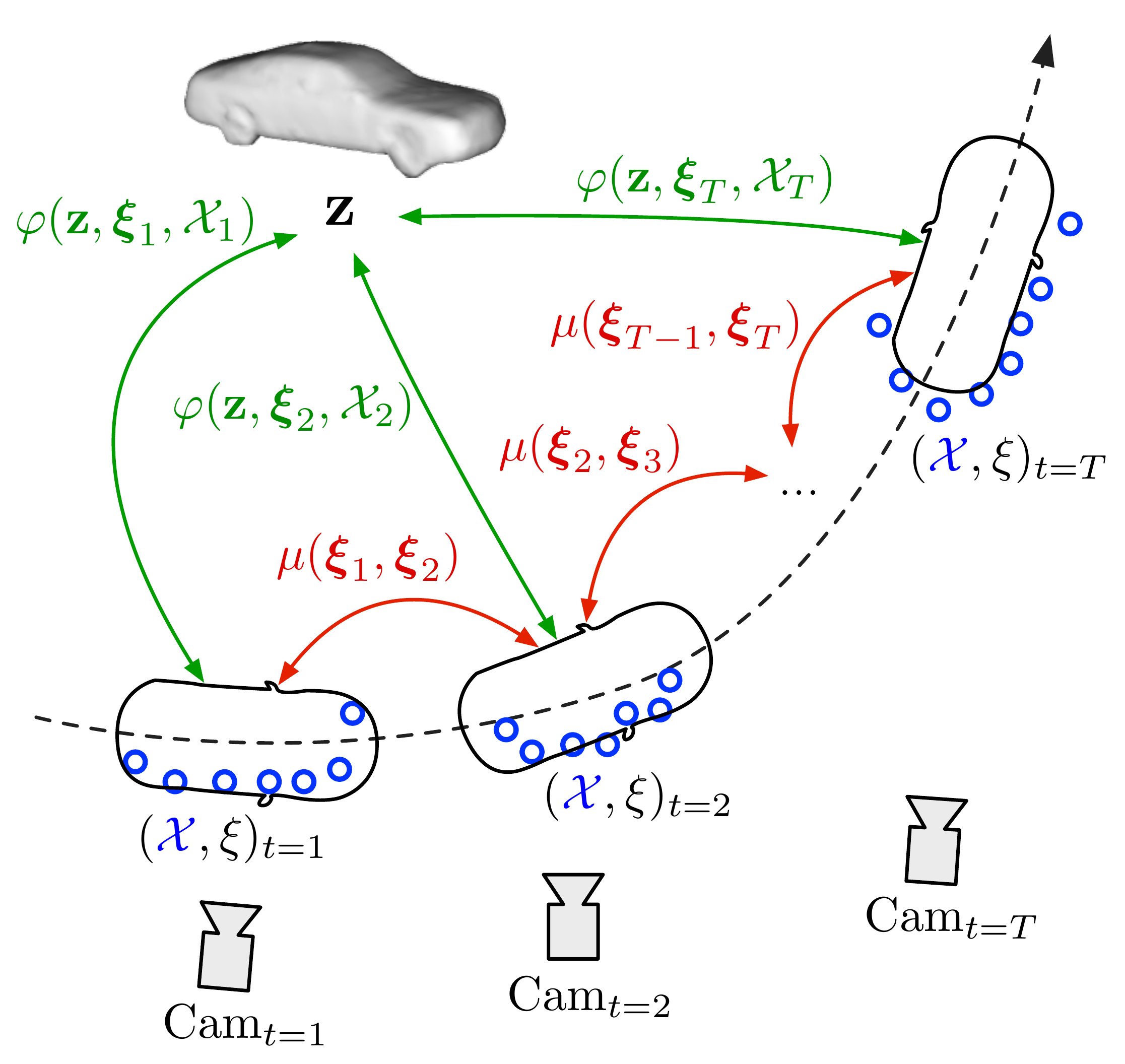}
\caption{\textbf{Illustration of Optimization.}
{\color[rgb]{0,0,1}Blue}: Stereo depth observations $\mathcal{X}$ of a tracked vehicle observed at different time steps $t$.
{\color[rgb]{1,0,0}Red}: Motion potentials enforcing consistent poses $\pose$ between successive frames.
{\color[rgb]{0,0.6,0}Green}: Shape potentials ensuring a constant shape $\shape$ of the tracked object along the track.}
\label{fig:optimization_illustration}
\end{figure}

We maximize the posterior probability introduced in the previous section by minimizing its negative logarithm.
This results in the following cost function:
\begin{equation}\label{eq:costfunction}
	\text{E}(\shape, \pose) =
	\frac{1}{T} \underset{t}{\sum}  [
	\underset{\text{data term}}{\underbrace{\varphi(\shape,\pose_t,\points_t)}}
	+ \,
	\underset{\text{motion term}}{\underbrace{\mu(\pose_t, \pose_{t-1}) }} \,
	 ] \, + 
	\underset{\text{shape prior}}{\underbrace{\kappa(\shape)}}
\end{equation}
In the following, we will explain the individual terms of the energy function in detail.
\reffig{optimization_illustration} shows an illustration of all the energy terms.

\subsubsection{Motion Term}
The motion term $\mu$ minimizes the difference between
the current pose $\pose_t$ and the predicted pose $g(\pose_{t-1})$
from the previous pose $\pose_{t-1}$ using a motion model~$g$:
\begin{equation}
\mu(\pose_t, \pose_{t-1}) = || \pose_t - g(\pose_{t-1})  ||^2_\mathbf{\Sigma} +   \underset{\text{ground plane prior}}{|| \underbrace{t_y - t_{y_{gp}}} ||^2_{\mathbf{\Sigma}_{gp}}}
\end{equation}
where the norm $|| \cdot ||^2_\mathbf{\Sigma} $ is defined as 
\begin{equation}
	|| \textbf{a} ||_\mathbf{\Sigma}^2 = \textbf{a}^\top \mathbf{\Sigma}^{-1} \textbf{a}.
\end{equation}
$\mathbf{\Sigma}$ is computed using first-order error-propagation of velocity noise through the motion model and adding a constant factor on the covariances for translation and rotation.
The supplementary material gives further details.
Additionally, we model a prior that keeps the vehicle attached to the ground plane.
In each frame, we estimate a ground plane as in \cite{Chen2015_3DOP},
where $t_{y_{gp}}$ is its elevation at the vehicle's position
and $\mathbf{\Sigma}_{gp}$ is the plane's uncertainty estimated from ground truth data.

We differentiate between three motion models for a tracked vehicle. They depend on the translation velocity $v$ and the angular velocity $\omega$ of the vehicle:
\begin{enumerate}
\item $\left|v\right| > \epsilon_v, \left|\omega\right| > \epsilon_{\omega}$: the vehicle takes a turn.
\item $\left|\omega\right| < \epsilon_{\omega}$: the vehicle drives straight on.
\item $\left|v\right| < \epsilon_v$: the vehicle is standing i.e. $g(\pose) = \pose$.
\end{enumerate}
where $\epsilon_v$ and $\epsilon_\omega$ are small threshold velocities. Section 3.4 describes the initialization of the velocities $v$ and $\omega$.

In the first case, we use a circular-arc motion model:
\begin{equation}
g(\pose) = \pose +
\begin{bmatrix}
 - \frac{v}{\omega} \, cos \theta + \frac{v}{\omega} \, cos(\theta + \omega \, \delta t) \\
0 \\
 + \frac{v}{\omega} \, sin \theta - \frac{v}{\omega} \, sin(\theta + \omega \, \delta t) \\
 \omega \, \delta t \\
 0 \\
 0
 \label{motion_model}
\end{bmatrix}
\end{equation}
In the second case, the vehicle moves on a straight line:
\begin{equation}
g(\pose) = \pose + v \, \delta t
\begin{bmatrix}
 sin \theta &
0 &
 -cos \theta&
 0 &
 0 &
 0
 \label{motion_model}
\end{bmatrix}^{\top}
\end{equation}
Note that the motion model assumes constant-position along the vertical y-direction.
Instead we model a prior term that keeps the tracked objects aligned to the ground plane.

\subsubsection{Data Term}

The data term $ \varphi$ measures the distance between the depth observations $\mathcal{X}$ and a vehicles' surface which is given implicitly by the zero level-set of the SDF~$\phi_{\shape}$:
\begin{equation}
 \varphi(\shape, \pose, \points) = \frac{1}{N} \underset{{\mathbf{x} \in \mathcal{X}}}{\sum} \rho \left( \frac{\phi_{\shape}(\textbf{T}_{\pose} \, \mathbf{x})}{\sigma_{d_\mathbf{x}}} \right),
\end{equation}
where $N$ is the number of elements in $\points$, $\textbf{T}_{\pose}$ the transformation matrix representing the estimated pose $\pose$, and $\rho$ the robust Huber-norm.

Further, we model the depth uncertainty~$\sigma_{d_{\mathbf{x}}}$ of each depth observation $\mathbf{x}$.
The uncertainty of $\mathbf{x}$ depends on its depth $d$ which is related to the estimated disparity $\delta$:
$d = \frac{b \, f}{\delta}.$
Using first-order error-propagation, we approximate the standard deviation $\sigma_d$ of the depth with
$\sigma_d = \frac{d^2 \, \sigma_{\delta}}{b \, f}$
where $\sigma_{\delta}$ is the standard deviation of $\delta$,
$b$ is the baseline of the stereo setup, and $f$ is the focal length of the cameras. Both $f$ and $b$ are known from the calibrated stereo camera setup.
Throughout our experiments we assume $\sigma_d$\,=\,1 pixel.

\subsubsection{Shape Prior}

As in~\cite{EngelmannGCPR16}, we use the shape prior term
\begin{equation}
\kappa( \shape ) = \overset{R}{\underset{i=1}{\sum}} \left( \frac{z_i}{\sigma_i} \right) ^2
\end{equation}
where $\sigma_i^2$ is the eigenvalue of the j-th principal component.
This term regularizes the shape parameters, i.e it penalizes too much deviation from the mean shape.

\begin{figure}[t]
\centering
\includegraphics[trim=17 0 17 0, clip, width=1.0\linewidth]{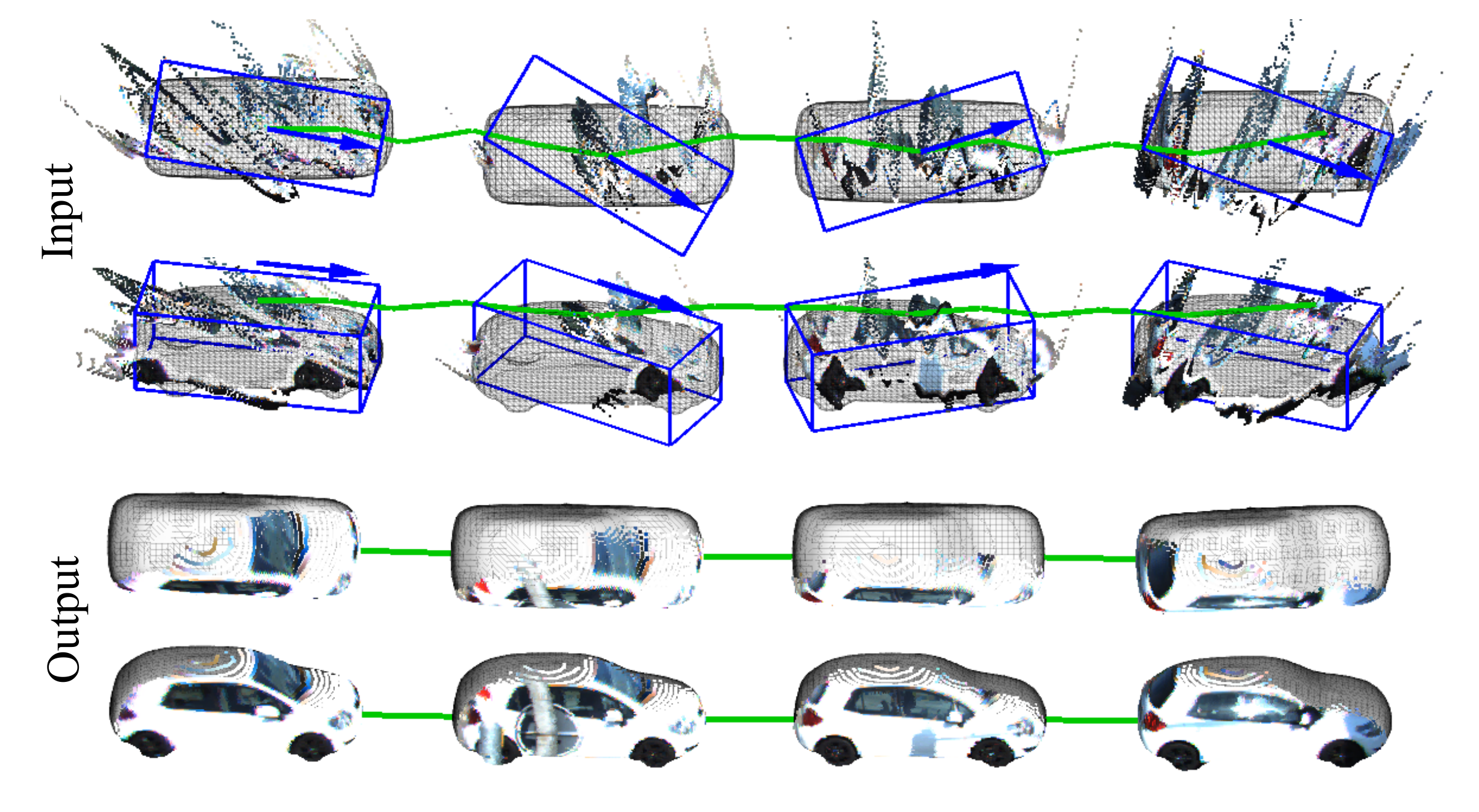}
\caption{\textbf{Noisy Input and Refined Output.} Top: Input of our method. Stereo depth observations of a vehicle, tracked over 20 frames. To avoid clutter we plot only four detections. Bounding boxes of 3DOP detections are shown in blue. The trajectory is shown in green. The wireframe shows our shape and pose initialization as described in Section 3.4. Bottom: Reconstructed shapes are aligned to the initial stereo observations. Color images are projected onto the shapes to show correct poses and shape. The smoothed trajectory is shown in green.}
\label{fig:quali1}
\end{figure}

\begin{figure*}[bt!]
\centering
\includegraphics[trim=0 450 0 0, clip, width=1.0\linewidth]{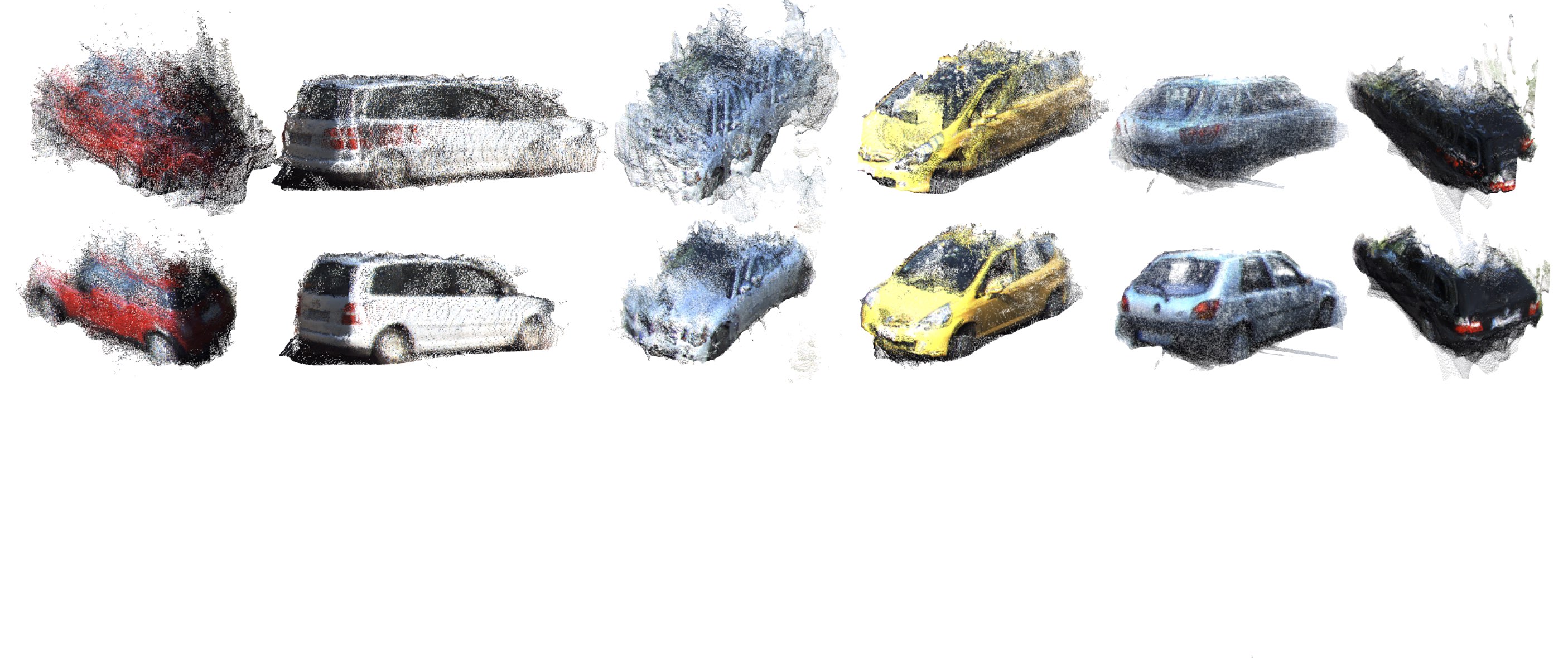}
\caption{\textbf{Performance of Pose Estimation.} The point clouds in the upper row show superimposed stereo depth observations $\points$ of a track based on their \textit{initial} poses $\pose$. The lower row shows the same point clouds but after performing our optimization.}
\label{fig:quali2}
\end{figure*}

\subsection{Optimization Details}
For each track, we initialize the shape of the tracked object to the mean shape.
The stereo depth observations $\points$ of an object in each frame are initialized with the stereo reconstructions that project into the 2D bounding box of the 3DOP detection.
We remove all observations (i.e. 3D points) below the estimated road plane and only keep those that lie within 3 m of the detected 3D position.
To find an initialization for the velocities we determine the median of the velocities $v$ and 
$\omega$ between successive detections along the track.
The translation $\textbf{t}\in \mathbb{R}^3$ is initialized to the position of the 3DOP detection projected onto the ground plane.
If a vehicle is static (see Sec. 3.3.1), we initialize its rotation to the 3DOP detection.
If it is moving, we align the forward heading of the vehicle with the segment connecting the first and last detection of the track.

We minimize the non-linear least-squares problem resulting from our cost function in \refeq{costfunction} with the Levenberg-Marquardt method for which we employ the Ceres solver~\cite{ceres-solver}.
The association of stereo reconstructions to vehicles through the 3DOP detections is sensitive to the accuracy of the detection.
In order to improve these associations, we use an alternating association and maximization procedure that corresponds to hard expectation-maximization:
After optimizing the cost function with the current association, we reassociate stereo depth observations using the current trajectory estimate instead of the 3DOP poses.
Consequently, we reoptimize the cost function with the new associations.
This procedure is repeated until convergence, whereas we observed that a single alternation suffices in our experiments.
In the initial pass, we parametrize uncertainty in the depth observations higher in order to account for the uncertain associations.

\section{Experimental Evaluation}
This section provides qualitative and quantitative results of our approach compared to several baseline methods.
We evaluated our method on the KITTI dataset \cite{Geiger2012CVPR}.
Specifically, we used the stereo 2015 multi-view extension.
It consists of 200 short video sequences of 21 frames each and shows dynamic vehicles in urban street scenes.
The middle frame of each sequence contains dense disparity annotations.
We used those to generate ground truth 3D points and compared them with our reconstructed shape.
For a subset of the sequences, ground truth pose annotations are available. We used those to evaluate our pose estimation.
For the quantitative evaluation we want to abstract from true negatives of the detector and incorrect associations from tracker.
Thus, we only consider sequences in which the densely depth annotated vehicles were successfully tracked.
The supplementary material lists the evaluated sequences.

\subsection{Baseline Methods}
For our evaluation, we leverage multiple baseline methods:
libELAS \cite{GeigerACCV10} which is a fast disparity estimation method,
SPS-Stereo \cite{YamaguchiECCV14} which also estimates disparities under the assumption of a locally piece-wise planar world,
our previous method \cite{EngelmannGCPR16} which makes use of shape priors to perform shape reconstruction and pose estimation on single frames,
and finally 3DOP \cite{Chen2015_3DOP} which is a state-of-the-art object detector.
We use the first three methods to compare the quality of our shape reconstruction and the two last to evaluate the pose estimation accuracy of our method.

\begin{table}[t]
\centering
\footnotesize
\begin{tabular}{lccc}
\toprule
Method 							& Accuracy 		& Completeness 	 		& \textbf{F1 score} \\
\midrule
libELAS \cite{GeigerACCV10} 			& 65.17 \% 		& \textbf{79.56} \% 			& 71.65 \% \\
SPS-St. \cite{YamaguchiECCV14} 		& 67.36 \%		& 70.93 \% 		 		& 69.10 \% \\
\cite{EngelmannGCPR16} (libELAS) 		& 75.08 \%		& 73.70 \% 		 		& 74.38 \% \\
\cite{EngelmannGCPR16} (SPS-St.) 		& 74.45 \% 		& 69.91 \% 				& 72.11 \% \\
Ours (libELAS) 						& \textbf{79.59} \%	& 74.79 \% 		 		& \textbf{77.11} \% \\
Ours (SPS-St.) 	 					& 77.52 \%		& 71.94 \% 		 		& 74.62 \% \\
\bottomrule
\\
\end{tabular}
\caption{\textbf{Quantitative Shape Evaluation Results.} This table shows the performance of our method compared to baseline methods with a distance threshold $\tau$\,=\,0.2 m.
In brackets we denote the stereo depth estimation methods used to initialize \cite{EngelmannGCPR16} and ours.
}
\label{tab:shape_eval}
\end{table}

\subsection{Shape Evaluation}
We analyze how well our shape model is able to reconstruct the surface of an observed vehicle.
Similar to \cite{ZhouICCV15}, we measure the performance in terms of completeness, accuracy and $F_1$ score.
The completeness is computed as the percentage of 3D ground truth points for which at least one reconstructed 3D point is within the distance of a threshold value $\tau$, the accuracy as the percentage of reconstructed 3D points for which at least one ground truth 3D point is within the distance of $\tau$. The F1 score is calculated as
\begin{equation}
F_1 = 2 \cdot \frac{\text{completeness} \cdot \text{accuracy}}{\text{completeness} + \text{accuracy}}.
\end{equation}
It trades off completeness and accuracy in a single evaluation measure.
Note that our estimated shape does not explicitly return reconstructed 3D points. We construct them indirectly by projecting the shape into the camera frame and reading out the depth buffer at each pixel that overlaps with the ground truth segment. 
The quantitative results are shown in \reftab{shape_eval} for $\tau$\,=\,0.2 m evaluated at all ground truth 3D points originating from the surface of a vehicle.
As shown by the experiments, our method outperforms the baseline methods in terms of accuracy for both depth initializations libELAS and SPS-Stereo.
Note that libELAS and SPS-Stereo are dense disparity estimation approaches operating on image data such that the depth is estimated for each image pixel, but the methods do not provide an instance segmentation like our method.
Since our instance segmentation might not fully overlap with the ground truth segment, our method yields moderately less completeness in this evaluation.
Also pose inaccuracies influence the results.
Furthermore, in this kind of evaluation, our method cannot benefit from its ability to predict full 3D shape for which no ground truth is available.
In the F1 score, which combines accuracy and completeness in a single measure, our approach consistently outperforms the other methods.

\begin{figure*}[t]
\centering
\includegraphics[trim=0 13 0 0, clip, width=1.0\linewidth]{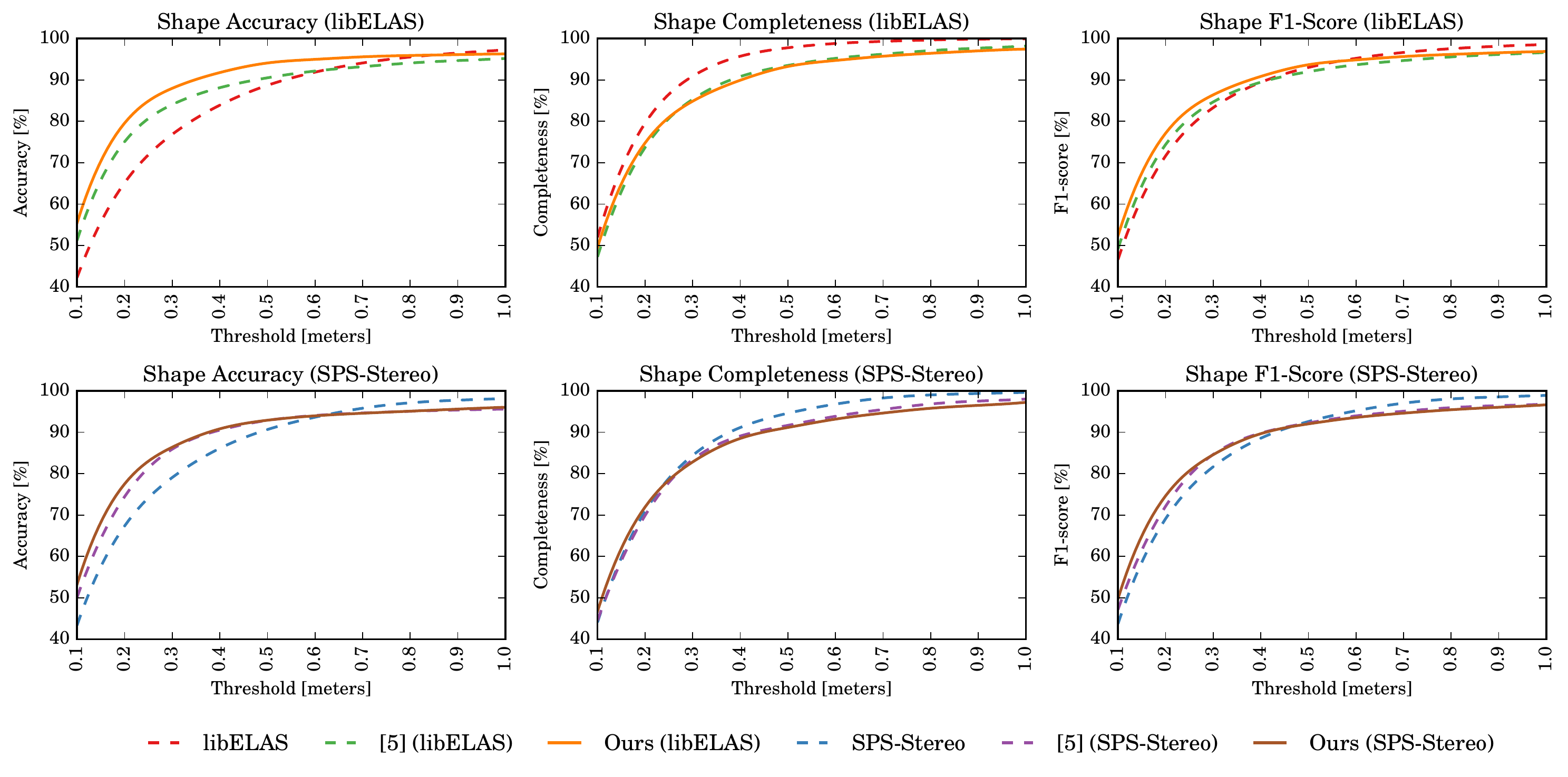}
\caption{\textbf{Quantitative Shape Evaluation Results.} Performance of our method compared to baseline methods with respect to completeness, accuracy and F1 score using a varying evaluation distance $\tau$. Top: Methods initialized with/or disparity maps from libELAS. Top: Methods initialized with/or disparity maps from SPS-Stereo.}
\label{fig:shape_eval}
\end{figure*}

Results for varying values of distance threshold $\tau$ are given in \reffig{shape_eval}. While accuracy and completeness scores drop for all methods with smaller distance threshold, the relative gain in accuracy compared to the baseline methods becomes more important and completeness achieves comparable values. This shows that the depth estimation of our method, although less complete, is more accurate.

\begin{figure}[tb]
\centering
\includegraphics[trim=0 20 0 20, clip, width=0.8\linewidth]{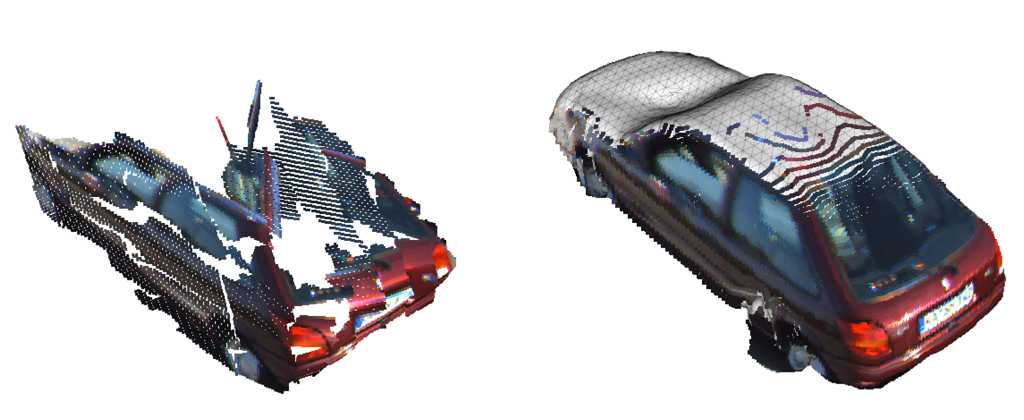}
\caption{\textbf{Model Completion.} Our method fills in unobserved areas and returns smoothed surfaces based on input stereo depth observations (Left: SPS-Stereo input, Right: Our output).}
\label{fig:model_completion}
\end{figure}

Furthermore, our model is able to reconstruct unseen or occluded parts of vehicles.
However, the qualitative evaluation hides this aspect due to lack of ground truth data. Instead, we show qualitative results in \reffig{model_completion}.

\begin{figure*}[t]
\centering
\includegraphics[trim=0 13 0 0, clip, width=0.97\linewidth]{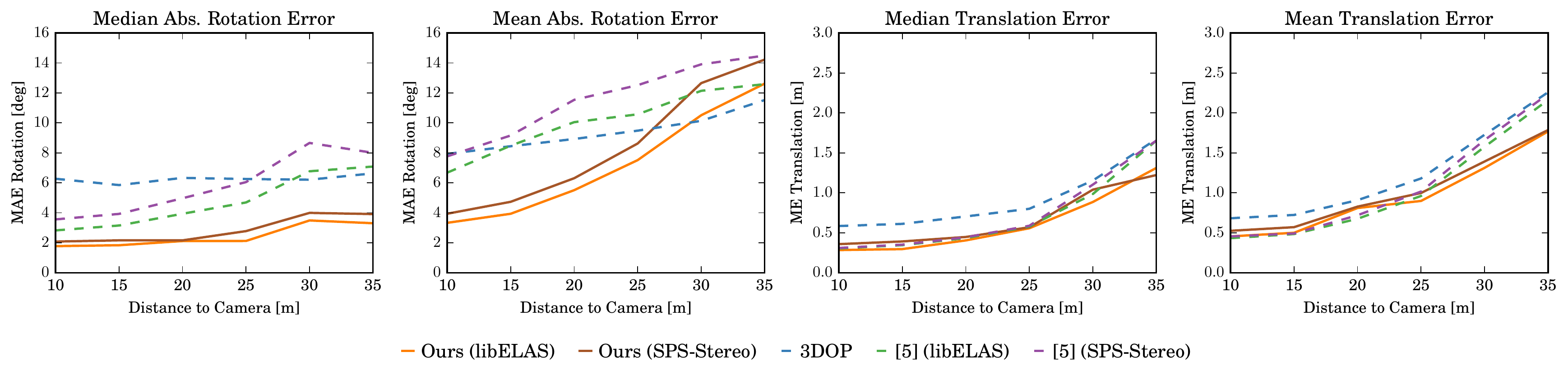}
\caption{\textbf{Quantitative Pose Evaluation.} Error of our pose estimation compared to multiple baselines for rotation (top) and translation (bottom). In addition to mean values (right), we also provide median values (left) which are less sensitive to outliers. Outliers can occur due to wrong tracking associations.}
\label{fig:eval_pose}
\end{figure*}

\subsection{Pose Evaluation}
To evaluate the pose estimation, we extracted ground truth pose annotations from KITTI's "raw data" section.
We evaluate the estimated pose for rotation and translation separately.
For the rotation, we evaluate the absolute rotation difference between ground truth and estimated rotation.
For the translation, we employ the Euclidean distance as error measure.
In \reffig{eval_pose}, we report results for varying distances between the tracked object and the camera.
At each distance step we consider a depth-window of 20 m.
Additionally to the mean errors, we also give the median errors which are less sensitive to outliers.
As evidenced by our plots, we largely outperform the baselines in terms of rotation and produce reasonable improvements in translation especially for large distances to the camera.

\setlength{\fboxsep}{0pt}%
\setlength{\fboxrule}{0.5pt}%

\begin{figure*}[!hbt]
\centering
\fbox{\includegraphics[trim=0 0 0 100, clip, width=0.495\linewidth]{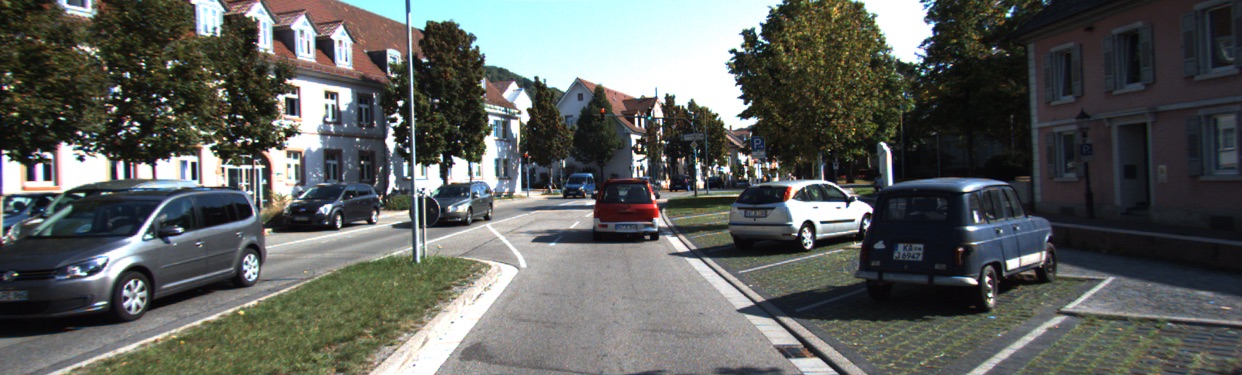}}
\fbox{\includegraphics[trim=0  0 0 100, clip, width=0.495\linewidth]{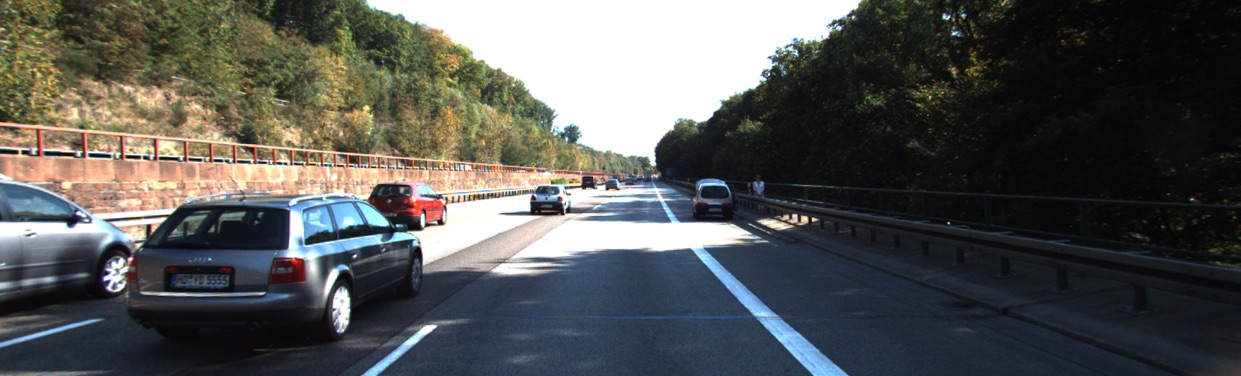}}\\
\vspace{2px}
\fbox{\includegraphics[trim=0 100 0 350, clip, width=0.495\linewidth]{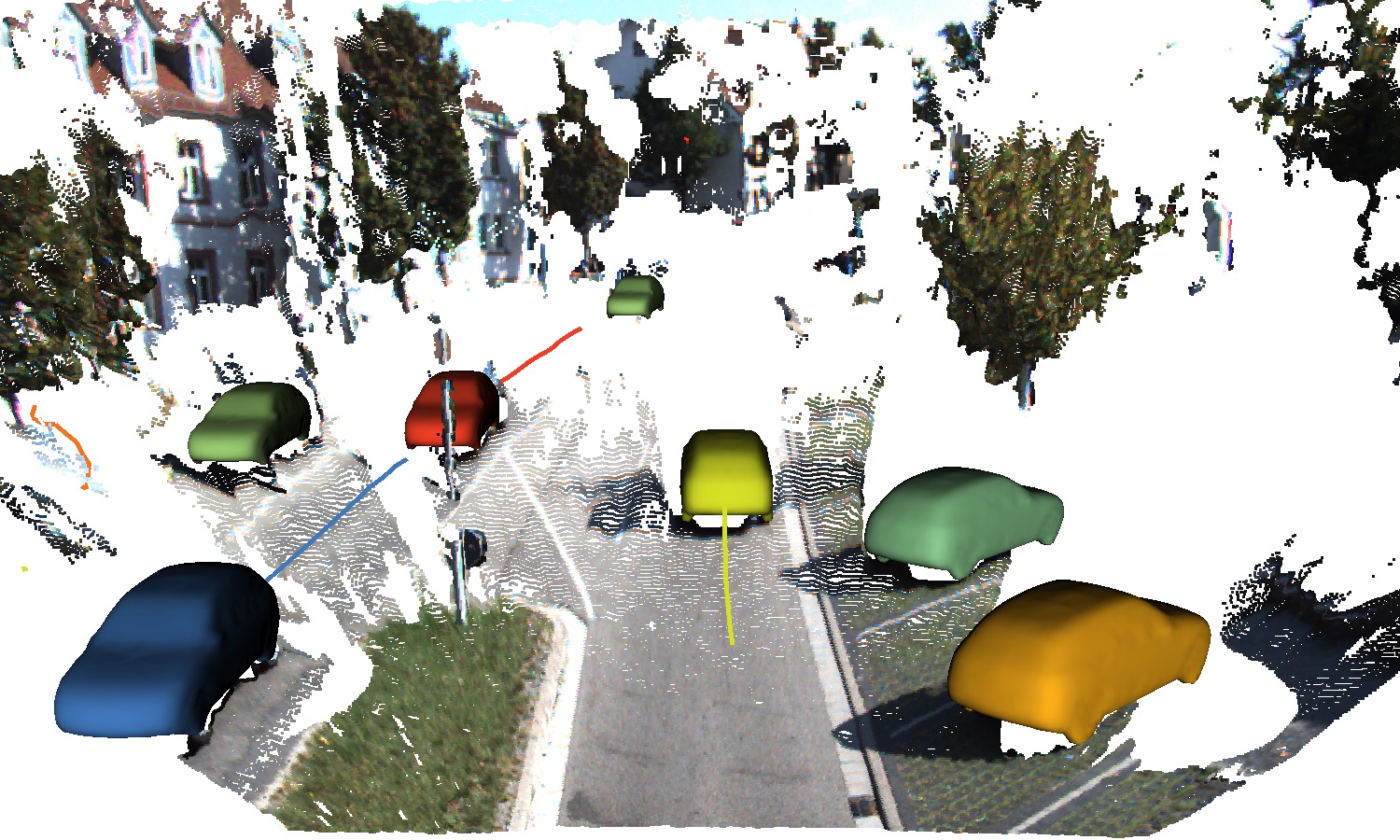}}
\fbox{\includegraphics[trim=0  300 0 150, clip, width=0.495\linewidth]{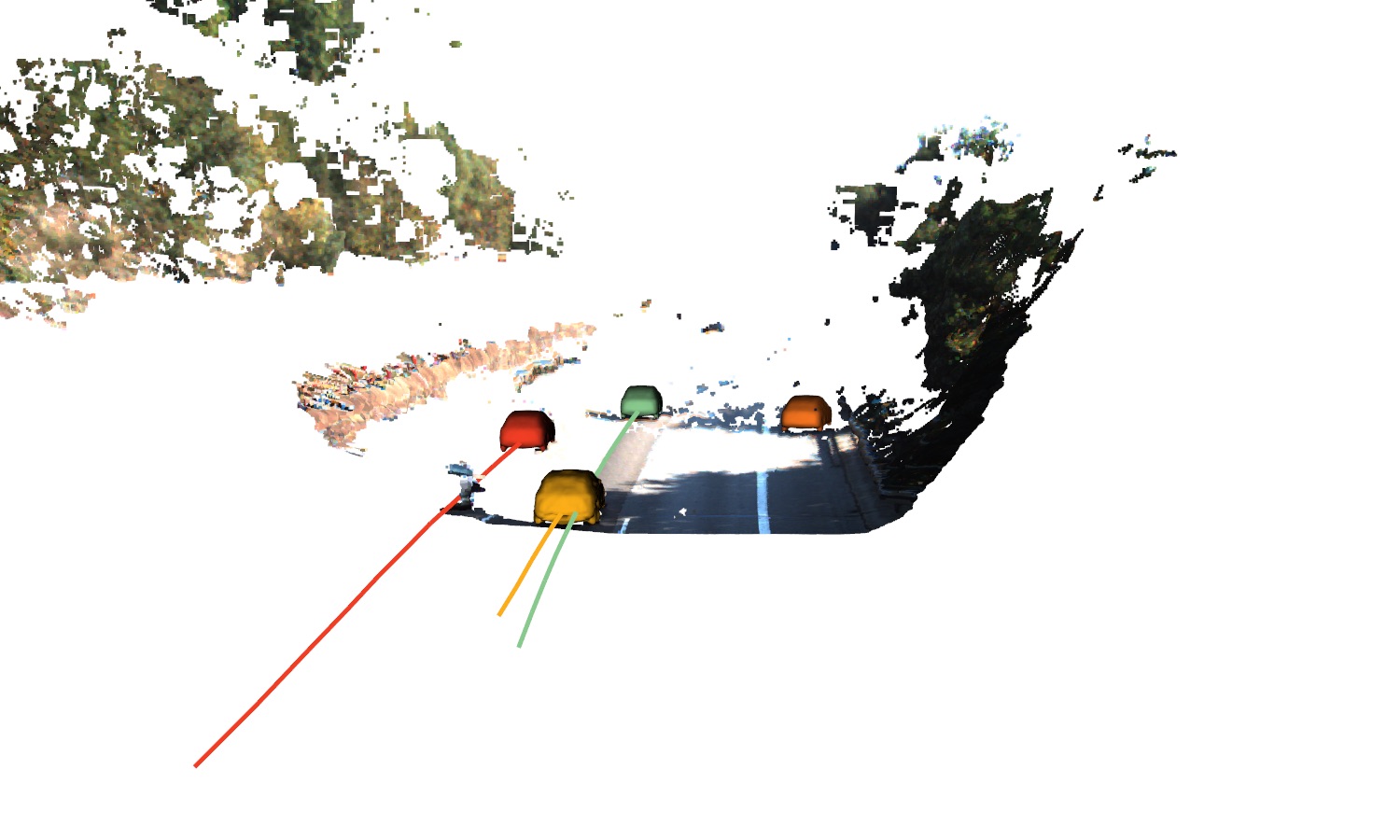}}\\
\vspace{2px}  
\fbox{\includegraphics[trim=0 0 0 100, clip, width=0.495\linewidth]{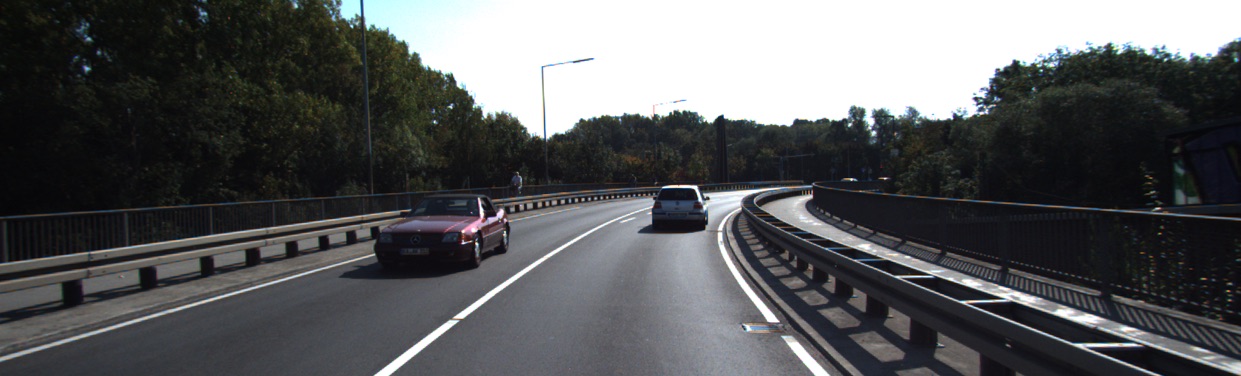}}
\fbox{\includegraphics[trim=0  0 0 100, clip, width=0.495\linewidth]{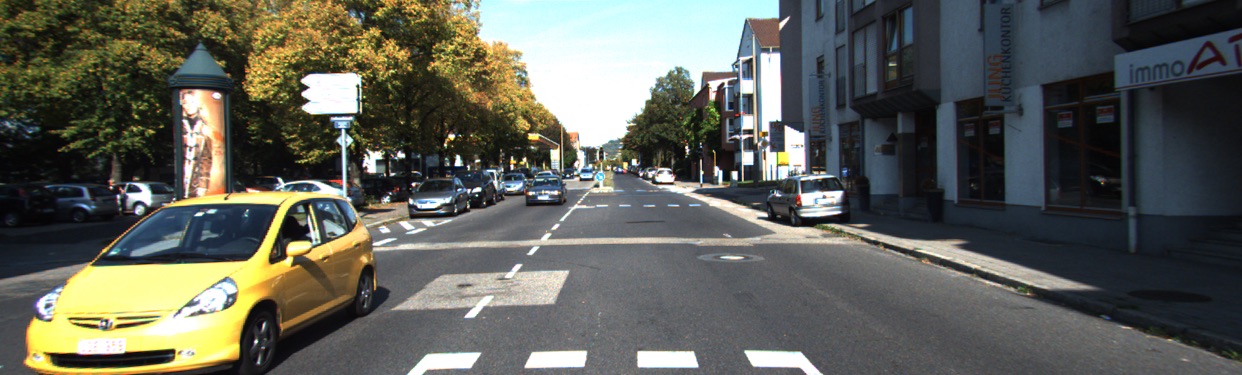}}\\
\vspace{2px}
\fbox{\includegraphics[trim=0 120 0 350, clip, width=0.495\linewidth]{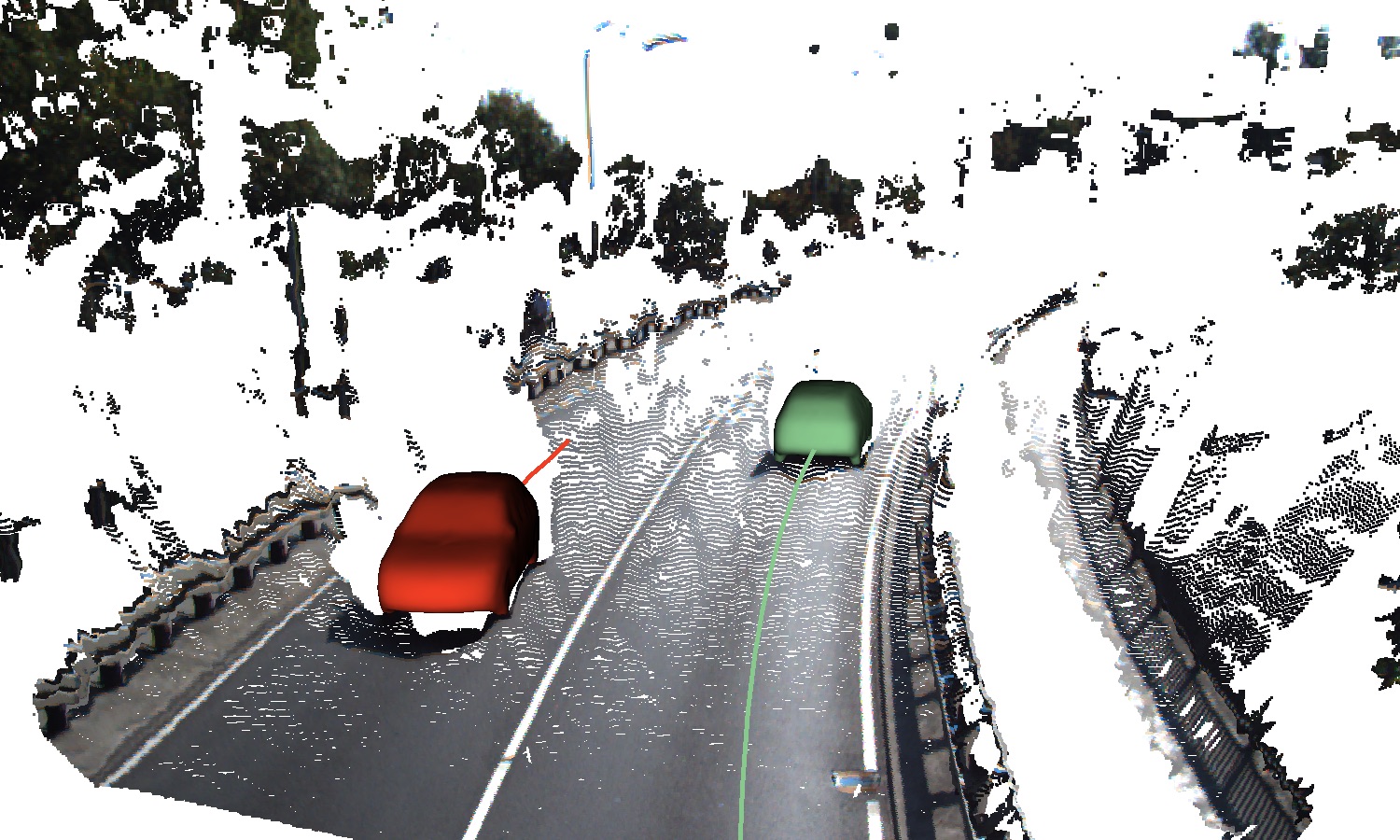}}
\fbox{\includegraphics[trim=0 120 0 350, clip, width=0.495\linewidth]{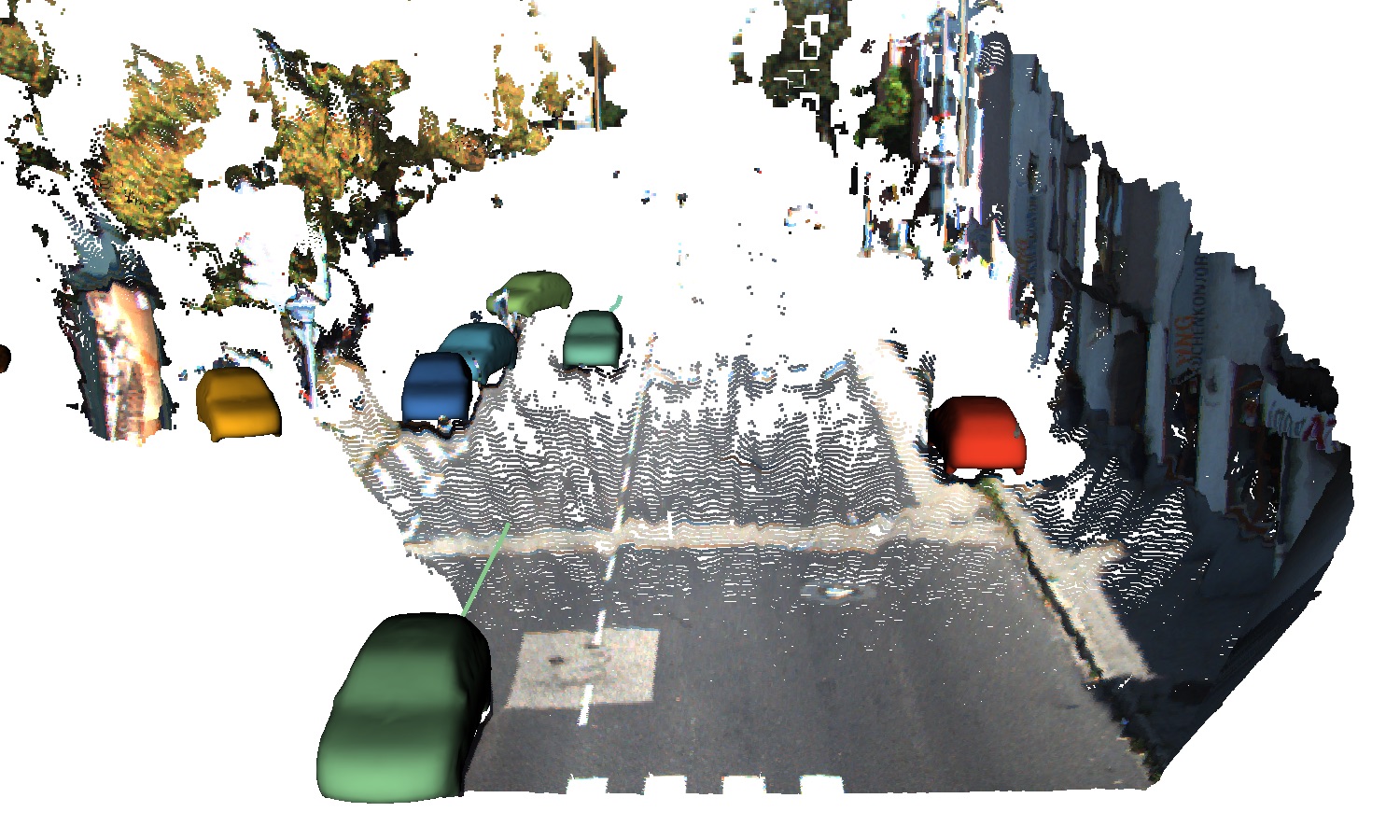}}\\
\vspace{2px}
\fbox{\includegraphics[trim=0 0 0 100, clip, width=0.495\linewidth]{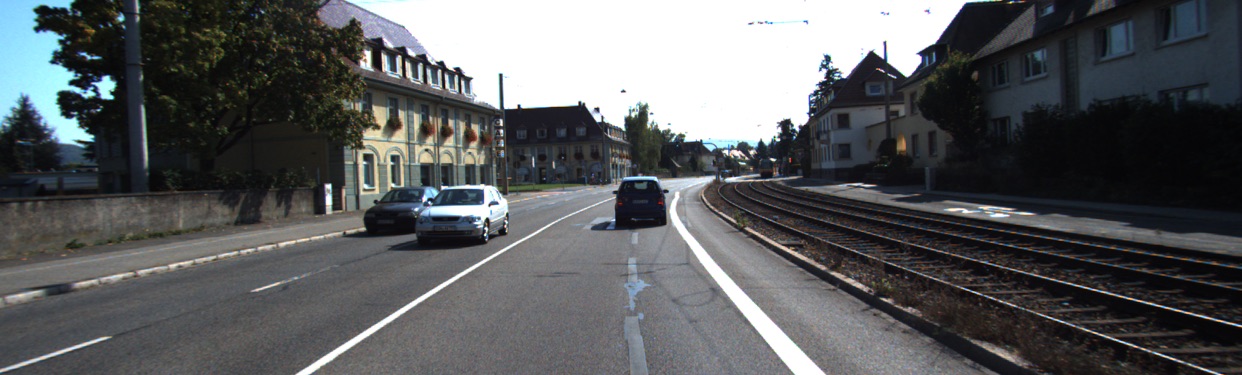}}
\fbox{\includegraphics[trim=0 0 0 100, clip, width=0.495\linewidth]{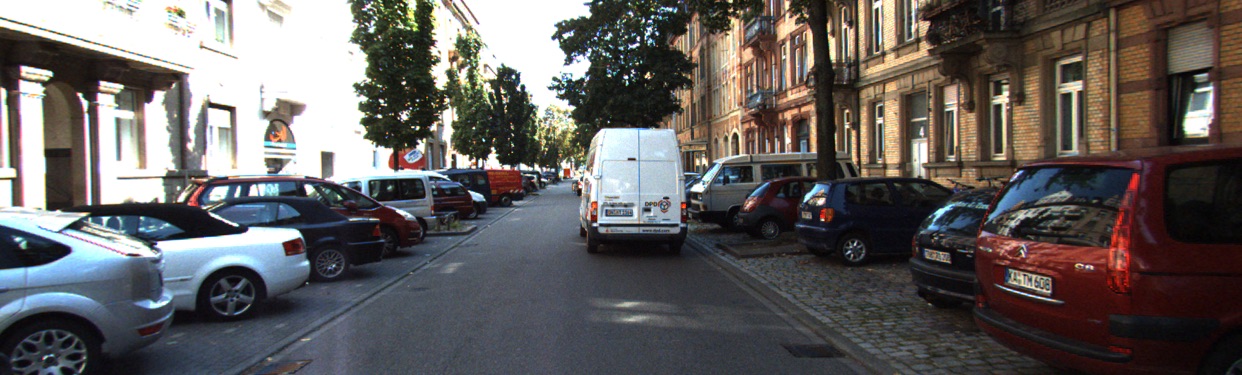}}\\
\vspace{2px}
\fbox{\includegraphics[trim=0 120 0 350, clip, width=0.495\linewidth]{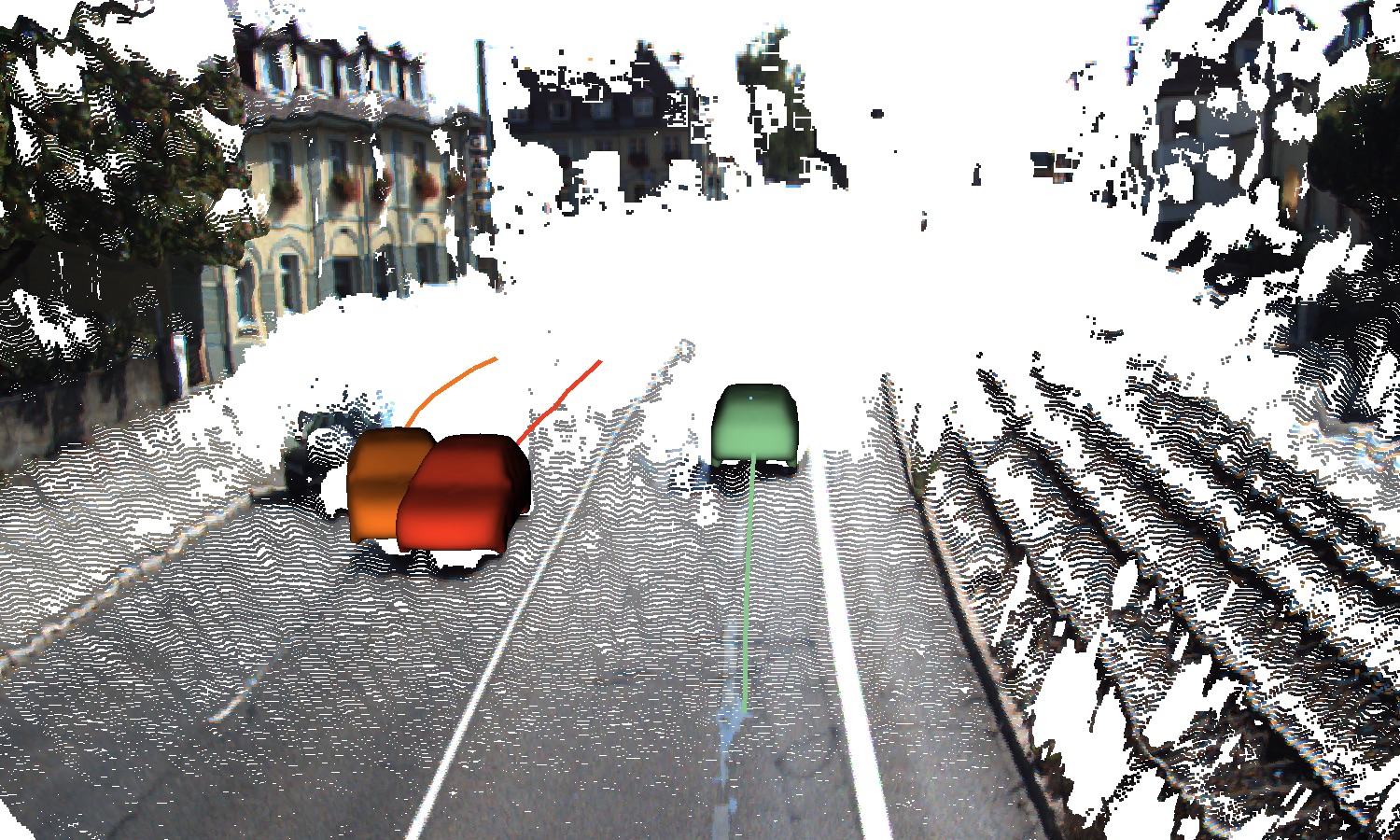}}
\fbox{\includegraphics[trim=0 120 0 350, clip, width=0.495\linewidth]{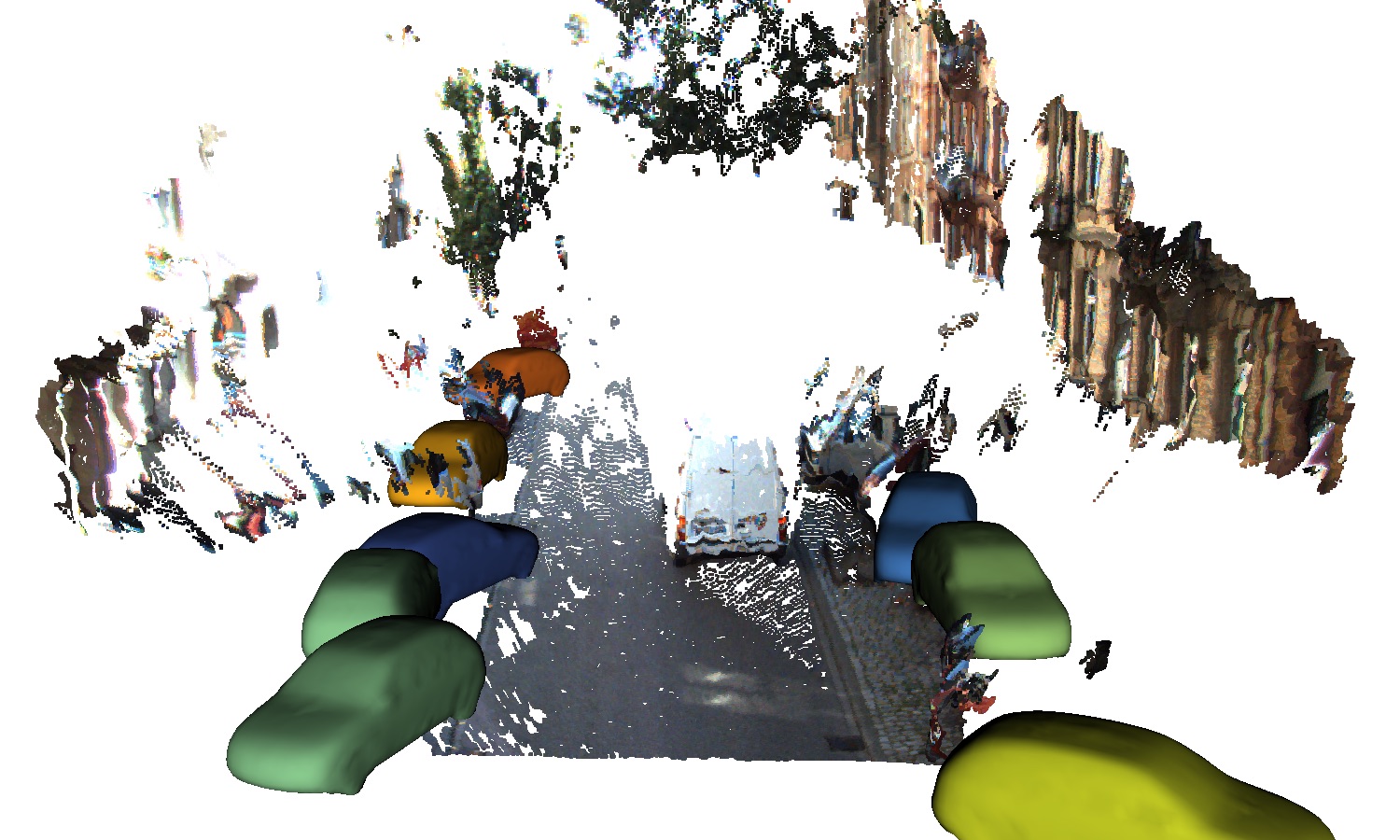}}

\caption{\textbf{Qualitative Results.} First two rows show exemplary good results of our method. Last Row: Failure cases.  Left: Initial tracklets of two vehicles are merged during the optimization due to noisy 3D detections. Right: Wrong initial detections and tracking associations that cannot be corrected by our method.     }
\label{fig:quali3}
\end{figure*}

\subsection{Qualitative Results}
We show qualitative results in Figures~\ref{fig:quali1}-\ref{fig:quali3}.
\reffig{quali1} shows input and initialization of our method in examples (top) and results after the optimization (bottom).
Note, that the second vehicle detection from the left is occluded by a street sign. 
Our method manages to estimate a smooth trajectory despite the occlusion. 
In~\reffig{quali2} we show qualitative results illustrating the performance of our pose estimation approach.
The overlaid point clouds from multiple frames are much more crisp using our pose estimate.
\reffig{model_completion} illustrates that our model is capable of reconstructing unobserved areas of vehicles through the shape embedding.
In~\reffig{quali3} we show multiple visual results alongside the color camera images.
The upper two rows demonstrate that our method recovers well object shape and smooth trajectories.
The lower row shows possible failure cases.
In the lower left example, the association of 3D detections to tracks fails and the tracks are merged.
On the lower right, many of the initial object detections, pose estimates and tracking associations are wrong such that our method fails.

\section{Conclusion}
In this paper, we present an approach for jointly estimating 3D the shape and the trajectory of vehicles in stereo image sequences of urban street scenes.
Our method uses an object tracker and a 3D object detector to generate initial trajectory hypotheses for shape alignment.
We model shapes in an embedding of SDFs which is learned from 3D CAD models.
We propose a probabilistic model of the stereo depth observations of the object shape within a track.
We assume the shape to remain constant across the object trajectory.
By this, the shape regularizes pose estimation across all frames and is optimized from all observations concurrently.
The formulation also includes a motion model which further regularizes the trajectory towards plausible motions.
The resulting non-linear least squares cost function for our probabilistic model is efficiently optimized using Levenberg-Marquardt.
In experiments we evaluate the accuracy of our approach in recovering shape and trajectory of the objects on the popular KITTI benchmark.
We demonstrate superior accuracy in shape reconstruction compared to state-of-the-art methods for stereo depth estimation and a single-frame shape alignment approach.
In future work, tight coupling of object tracking with our optimization scheme could be investigated for joint tracking, shape and trajectory estimation.
A further direction of research is to learn embeddings of multiple classes to model a larger variety of objects.
\\\\
\textbf{Acknowledgements}.
This work has been supported by ERC Starting Grant CV-SUPER (ERC-2012-StG-307432).

\balance

{\small
\bibliographystyle{ieee}
\bibliography{abbrev_short,egbib}
}

\end{document}